# 基于稀疏连接的层次化多核 *K*-Means 算法

王 雷[1,2] 杜 亮[1,2] 周 芃[3]
1 山西大学计算机与信息技术学院 太原 030006
2 山西大学大数据科学与产业研究院 太原 030006
3 安徽大学计算机科学与技术学院 合肥 230601
(575264909@qq.com)

**摘 要** 多核学习(Multiple Kernel Learning,MKL)的目标是寻找一个最优的一致性核函数。在层次化多核聚类算法(HMKC)中,通过从高维空间中对样本特征进行逐层提取的方式来实现最大化地保留有效信息,但是却忽略了层与层之间的信息交互。该模型中只有相邻层中对应的结点会进行信息交互,对于其他结点来说是孤立的,而采用全连接的方式又会削弱最终一致性矩阵的多样性。因此,文中提出了一种基于稀疏连接的层次化多核 *K*-Means 算法(Sparse Connectivity Hierarchical Multiple Kernel *K*-Means,SCHMKKM)。该算法通过稀疏率来控制分配矩阵以达到稀疏连接的效果,从而将层与层之间信息蒸馏得到的特征进行局部融合。最后,在多个数据集上进行聚类分析,并在实验中与全连接的层次化多核 *K*-Means 算法(FCHMKKM)进行实验对比,证明了具有更多差异性的信息融合有利于学习更好的一致性划分矩阵,并且稀疏连接的融合策略优于全连接的策略。

## Hierarchical Multiple Kernel *K*-Means Algorithm Based on Sparse Connectivity

WANG Lei[1,2],DU Liang[1,2] and ZHOU Peng[3]
1 College of Computer and Information Technology,Shanxi University,Taiyuan 030006,China
2 Institute of Big Data Science and Industry,Shanxi University,Taiyuan 030006,China
3 College of Computer Science and Technology,Anhui University,Hefei 230601,China

**Abstract** Multiple kernel learning(MKL) aims to find an optimal consistent kernel function. In the hierarchical multiple kernel clustering(HMKC) algorithm,the sample features are extracted layer by layer from high-dimensional space to maximize the retention of effective information,but the information interaction between layers is ignored. In this model,only the corresponding nodes in the adjacent layer will exchange information,but for other nodes,it is isolated,and if the full connection is adopted,the diversity of the final consistence matrix will be reduced. Therefore,this paper proposes a hierarchical multiple kernel *K*-Means (SCHMKKM) algorithm based on sparse connectivity,which controls the assignment matrix to achieve the effect of sparse connections through the sparsity rate,thereby locally fusing the features obtained by the distillation of information between layers. Finally,we perform cluster analysis on multiple data sets and compare it with the fully connected hierarchical multiple kernel *K*-Means(FCHMKKM) algorithm in experiment. Finally,it is proved that more discriminative information fusion is beneficial to learn a better consistent partition matrix,and the fusion strategy of sparse connection is better than the strategy of full connection.

**Keywords** Multiple kernel learning,Hierarchical multiple kernel clustering,Sparse connectivity,Fully connected,Information distillation,Local fusion

*K*-Means 算法是由 MacQueen[1] 于 20 世纪 60 年代提出,由于其可以高效地处理线性可分的大数据集,至今仍然是最受欢迎和最普遍使用的算法之一。但是在处理非线性复杂数据时,*K*-Means 算法往往表现不佳。为了弥补这一缺陷,核学习与 *K*-Means 算法相结合的算法相框架被提出,称作核 *K*-Means(KKM)[2] 算法。该算法将 *K*-Means 中度量样本间相似度的欧氏距离替换为非线性核函数,从而将非线性数据映射到高维特征空间实现线性可分。随着核学习的成功,基于核函数的聚类算法越来越受欢迎,代表性的核聚类方法有核 *K*-Means[2]、核模糊 C 均值聚类(KFCM)[3]、核谱聚类(KSC)[4]等。然而随着研究的深入,基于核函数的聚类算法在性能上很大程度会受到初始预定义核的影响,因此通过提

**到稿日期**:2022-04-24 **返修日期**:2022-10-27
**基金项目**:国家自然科学基金面上项目(61976129,62176001);山西省青年科技研究基金(201901D211168)
This work was supported by the National Natural Science Foundation of China(61976129,62176001) and Natural Science Foundation for Young Scientists of Shanxi Province,China(201901D211168).
**通信作者**:杜亮(duliang@sxu.edu.cn)

升核函数的质量来提升聚类性能成为了多核聚类算法的关键。在此期间许多有效的算法被提出,如最优核 $K$-Means 聚类(OKKC[5])、多核学习(MKL[6])、基于 $L_{21}$ 范数的鲁棒多核 $K$-Means 算法(RMKKM[7])等。也有应用于不同场景下的多核学习方法,如子空间多核聚类(KSSC[8],PMKSC[9])、层次化多核聚类(HMKC[10])、最优邻域的多核聚类算法(ONKC[11],ONALK[12])等。

在 HMKC 算法中,隐藏层中的结点只与其相邻层中对应的结点有信息交互,而与其相邻层中的其他结点则没有信息传递。受卷积神经网络[13]中神经元结点之间的信息交互和稀疏连接思想[14]的影响,本文提出了一种稀疏连接的层次化多核 $K$-Means 算法,试图探究层次化多核聚类中层与层之间的信息交互对聚类性能的影响以及层次化聚类中稀疏连接和全连接的区别。本文的主要贡献有以下 3 点:

(1)提出了一种基于稀疏连接的层次化 $K$-Means 算法(SCHMKKM),该算法将层次化多核聚类中相邻层之间不同的结点通过稀疏连接的策略实现了局部判别信息的融合。

(2)对比实验证明加入局部判别信息融合而学到的一致性划分矩阵可以提升聚类性能。

(3)对比实验证明稀疏连接策略相较于全连接策略在聚类性能上更优。

## 1　相关工作

### 1.1　多核 $K$-Means

对于数据集 $X\in\Re^{n\times d}$ 和一个核映射函数 $\phi(\cdot)$。核 $K$-Means 算法的目的是通过核函数将数据映射到新的特征空间,即 $\phi(\cdot):x\in X\ a\ f\in\Re^{d'}$,进而完成在高维特征空间中的线性可分。其中由核函数 $\phi(\cdot)$ 隐式定义的线性空间,被称为再生核希尔伯特空间(Reproducing Kernel Hilbert Space,RKHS),用 $f_\phi$ 表示。假设数据 $X$ 被划分为 $k$ 个簇 $C=\{C_1,C_2,\cdots,C_k\}$,则核 $K$-Means 可以定义为以下优化 $Z$ 和 $\mu$ 的模型:

$$\min_{Z,\mu}\sum_{i=1}^{n}\sum_{j=1}^{k}Z_{ij}\|\phi(x_i)-\mu_j\|_2^2\quad \text{s.t.}\ \sum_{j=1}^{k}Z_{ij}=1 \tag{1}$$

其中,$Z$ 为聚类分配矩阵,$Z_{ij}\in\{0,1\}^{n\times k}$ 表示是否将样本 $x_i$ 分配给簇 $C_j$。$\mu_j=1/n_{C_j}\sum_{i=1}^{n}Z_{ij}\phi(x_i)$ 表示簇 $C_j$ 的聚类中心,$n_{C_j}$ 表示属于簇 $C_j$ 的样本数目。由于 $Z$ 是离散的,使得式(1)求解变得困难,一种常见的做法是将 $Z$ 变松弛从而转换为一个新的变量 $H=ZL^{1/2}$,其中 $L=diag([n_{C_1}^{-1},n_{C_2}^{-1},\cdots,n_{C_k}^{-1}])$,故可以将式(1)转换为优化变量 $H$ 的形式:

$$\min_{H}\ \mathrm{tr}(K(\mathrm{I}_n-HH^{\mathrm T}))\quad \text{s.t.}\ H^{\mathrm T}H=\mathrm{I}_k \tag{2}$$

其中,$\mathrm{I}_n$ 为 $n\times n$ 的单位矩阵。多核 $K$-Means 是文献[6]中提出的一种通过权重 $w$ 将多个核函数融合为一个更好的一致性核函数 $K$ 的方法。形式如下:

$$\begin{aligned}&\min_{H,w}\sum_{p=1}^{m}w_p\,\mathrm{tr}(K_p(\mathrm{I}_n-HH^{\mathrm T}))\\&\text{s.t.}\ \ H^{\mathrm T}H=\mathrm{I}_k,\sum_{p=1}^{m}w_p=1,w_p\geqslant 0\end{aligned} \tag{3}$$

其中,一致性核函数 $K=\sum_{p=1}^{m}w_p^2K_p$。

### 1.2　稀疏连接

受神经科学中细胞只对一个视觉区域内一部分敏感而对其他部分视而不见的现象启发,稀疏连接成为卷积神经网络的一个重要思想:稀疏连接可以通过极少数的参数就可以达到与全连接相同的效果。

由于应用场景的不同,本文的研究与深度学习[15]存在以下差异:1)神经网络中的全连接层的作用之一是维度变换,尤其是将高维特征变换到低维以保留有用的数据信息,但是全连接层无法保持特征的空间结构,而核学习的目的是将原始数据的特征映射到高维空间中做内积然后探索数据的空间结构,让其在高维空间中变得线性可分;2)神经网络中的全连接层的参数占整个神经网络总参数的 80%,而最近的研究也证明,有过多的参数是冗余的[14]。

## 2　稀疏连接的层次化多核 $K$-Means 算法

图 1 给出了本文提出的 SCHMKKM 算法和相关算法的结构示意图。

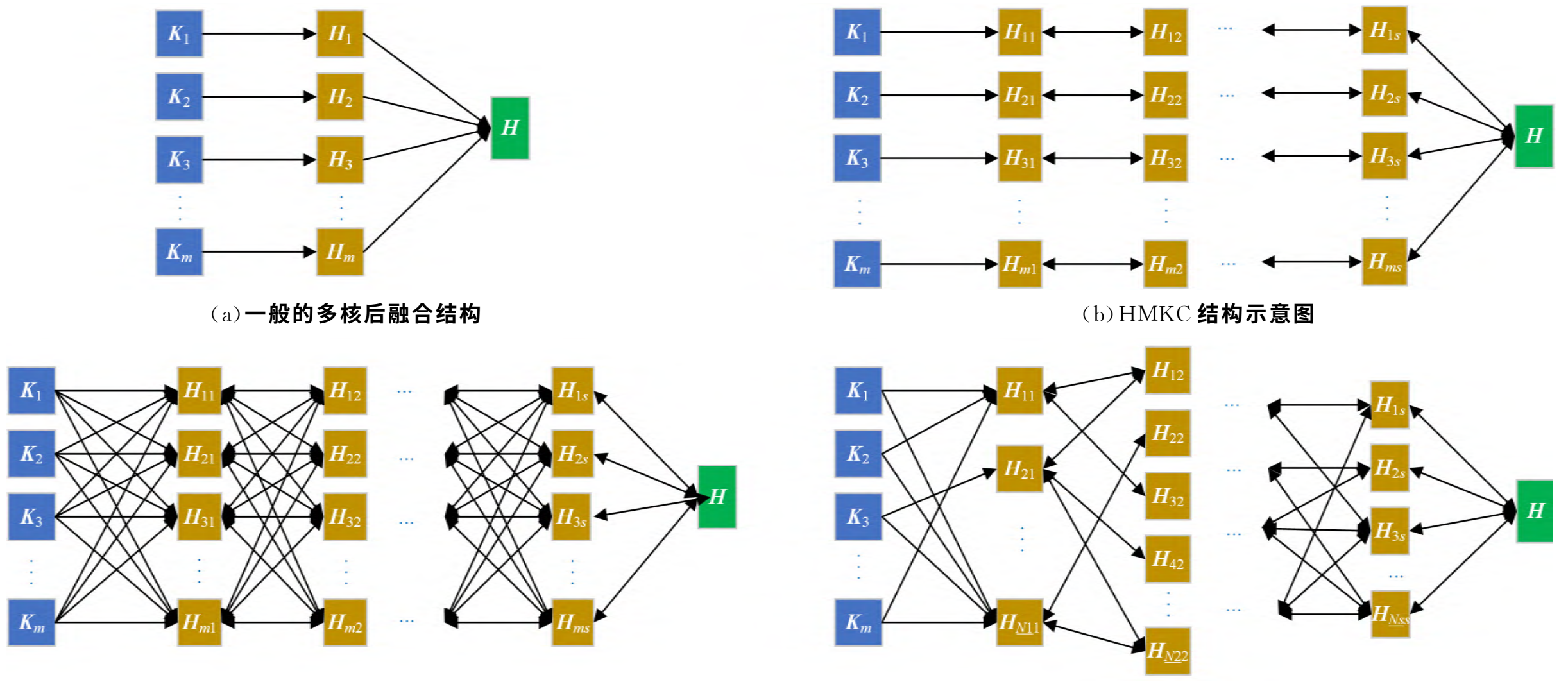

(a)一般的多核后融合结构　(b)HMKC 结构示意图　(c)FCHMKKM 结构示意图　(d)SCHMKKM 结构示意图

图 1　4 种不同后融合多核聚类算法的结构示意图

Fig. 1　Schematic diagram of four different late fusion multiple kernel clustering algorithms

图 1(a)为多核融合算法中常见的一种多核融合策略——后融合(Late Fusion, LF)。图 1(b)为 HMKC 算法的结构示意图,HMKC 是一种多层的多核融合算法,它与 LF 算法主要有两方面的区别。首先经典的 LF 算法通常采用一层结构,即先将多个 $n\times n$ 的核通过 KKM 算法转化为对应的多个 $n\times k$ 的划分矩阵,再将多个划分矩阵融合为一个一致性的划分矩阵,其中 $n\gg k$。在将数据从 $n$ 维空间提取至 $k$ 维空间的过程中造成了重要信息的丢失。而 HMKC 通过多层结构实现了逐步的信息提取,最大化地保留了有效信息。其次经典的 LF 算法没有中间划分矩阵与最终一致性矩阵的信息交互(图 1(a)中 $\boldsymbol{H}_i\rightarrow\boldsymbol{H}^*$ 为单向箭头),即没有"鸡生蛋,蛋生鸡"的过程。这使得多核融合成了一个静态的、单向的过程,而 HMKC 采用的多层动态的多核融合框架(图 1(b)中 $\boldsymbol{H}_i^{t-1}\rightarrow\boldsymbol{H}_i^t\rightarrow\boldsymbol{H}^*$ 均为双向箭头)也在相关实验中被证明了其相比原来的 LF 算法具有优越性。

HMKC 中每一层的结点都只和其相邻两层的同一线程的结点($\boldsymbol{H}_i^{t-1},\boldsymbol{H}_i^{t+1}$)有信息交互,而与不同线程的其他结点($\boldsymbol{H}_{j\neq i}^{t-1},\boldsymbol{H}_{j\neq i}^{t+1}$)没有信息交互。故本文提出一种稀疏连接的多核 $K$-Means 算法(见图1(d))来探索不同层之间结点之间信息交互对聚类性能的影响,并进一步对比了全连接(见图 1(c))与稀疏连接的区别。

### 2.1 构建模型

为了更好地理解 SCHMKKM 模型的架构,我们将其分为 3 个部分来论述。

第一部分:$\boldsymbol{K}\rightarrow\boldsymbol{H}_{N_1}^{(1)}$。第一层的第 $N_1$ 个划分矩阵 $\boldsymbol{H}_{N_1}^{(1)}$ 由 $m$ 个预定义核矩阵 $\boldsymbol{K}=\{\boldsymbol{K}_1,\boldsymbol{K}_2,\cdots,\boldsymbol{K}_P\}$ 通过 KKM 算法得到,从核矩阵到划分矩阵的相似性保持通过优化 $\boldsymbol{H}_i^{(1)}$ 和 $w$ 实现:

$$\begin{aligned}&\max_{\boldsymbol{H}_i,\boldsymbol{w}}\operatorname{tr}\Big(\boldsymbol{H}_i^{(1)\mathrm{T}}\Big(\sum_{p=1}^{m}\boldsymbol{dr}_{ip}^{(1)}\boldsymbol{w}_{ip}^{(1)}\boldsymbol{K}_p\Big)\boldsymbol{H}_i^{(1)}\Big)\\&\text{s.t. }\boldsymbol{H}_i^{(1)\mathrm{T}}\boldsymbol{H}_i^{(1)}=\mathbf{I}_{c_1},\boldsymbol{H}_i^{(1)}\in\Re^{n\times c_1}\\&\qquad\boldsymbol{w}_i^{(1)\mathrm{T}}\boldsymbol{w}_i^{(1)}=1\end{aligned}\tag{4}$$

其中,$\boldsymbol{dr}_i^{(1)}=\{\boldsymbol{dr}_{i1}^{(1)},\boldsymbol{dr}_{i2}^{(1)},\cdots,\boldsymbol{dr}_{im}^{(1)}\}$,$\boldsymbol{dr}_{ip}^{(1)}\in\{0,1\}$ 为 Dropout 矩阵,当选择核矩阵 $\boldsymbol{K}_i$ 时 $\boldsymbol{dr}_{ip}^{(1)}=1$,否则为 0。每一层的 Dropout 矩阵由总体的稀疏率(Sparsity Ratio, SR)决定,稀疏率越高,连接越少(0 为全连接)。$\boldsymbol{w}_{im}^{(1)}=\{\boldsymbol{w}_{i1}^{(1)},\cdots,\boldsymbol{w}_{im}^{(1)}\}$,$\boldsymbol{w}_{ip}^{(1)}\in[0,1]$ 为核矩阵 $\boldsymbol{K}_i$ 的权重矩阵。$c_1$ 表示第一层核 $K$-Means 算法的聚类数。

第二部分:$\boldsymbol{H}_{N_{t-1}}^{(t-1)}\rightarrow\boldsymbol{H}_{N_t}^{(t)}$。中间层的第 $N_t$ 个划分矩阵 $\boldsymbol{H}_{N_t}^{(t)}$ 由前一层的 $N_{t-1}$ 个矩阵 $\boldsymbol{H}_{N_{t-1}}^{(t-1)}$ 稀疏连接构成,它们的相似性保持通过以下模型优化 $\boldsymbol{H}_i^{(t)}$ 和 $w$ 实现:

$$\begin{aligned}&\max_{\boldsymbol{H}_i,\boldsymbol{w}}\operatorname{tr}\Big(\boldsymbol{H}_i^{(t)\mathrm{T}}\Big(\sum_{j=1}^{N_{t-1}}\boldsymbol{dr}_{ij}^{(t-1)}\boldsymbol{w}_{ij}^{(t-1)}\boldsymbol{H}_j^{(t-1)}\boldsymbol{H}_j^{(t-1)\mathrm{T}}\Big)\boldsymbol{H}_i^{(t)}\Big)\\&\text{s.t. }\boldsymbol{H}_i^{(t)\mathrm{T}}\boldsymbol{H}_i^{(t)}=\mathbf{I}_{c_t},\boldsymbol{w}_i^{(t-1)\mathrm{T}}\boldsymbol{w}_i^{(t-1)}=1,\boldsymbol{H}_i^{(t)}\in\Re^{n\times c_t}\end{aligned}\tag{5}$$

其中,$c_t$ 表示第 $t$ 层的聚类数,$N_t$ 表示第 $t$ 层的划分矩阵总数。

第三部分:$\boldsymbol{H}_{N_s}^{(s)}\rightarrow\boldsymbol{H}$。最后的一致性核矩阵 $\boldsymbol{H}$ 由第 $s$ 层的 $N_s$ 个划分矩阵线性融合得到,优化目标 $\boldsymbol{H}$ 和 $\boldsymbol{\beta}$ 的模型如下:

$$\begin{aligned}&\max_{\boldsymbol{H},\boldsymbol{\beta}}\sum_{i=1}^{N_s}\boldsymbol{\beta}_i\operatorname{tr}(\boldsymbol{H}^{\mathrm{T}}\boldsymbol{H}_i^{(s)}\boldsymbol{H}_i^{(s)\mathrm{T}}\boldsymbol{H})\\&\text{s.t. }\boldsymbol{H}^{\mathrm{T}}\boldsymbol{H}=\mathbf{I}_k,\boldsymbol{\beta}^{\mathrm{T}}\boldsymbol{\beta}=1\end{aligned}\tag{6}$$

其中,$k$ 表示最终聚类数,$s$ 表示中间过渡层的总数。

将式(4)-式(6)整合到统一框架中,即为 SCHMKKM 算法的优化模型:

$$\begin{aligned}&\max\sum_{i=1}^{N_1}\boldsymbol{\gamma}_i^{(1)}\operatorname{tr}\Big(\boldsymbol{H}_i^{(1)\mathrm{T}}\Big(\sum_{p=1}^{m}\boldsymbol{dr}_{ip}^{(1)}\boldsymbol{w}_{ip}^{(1)}\boldsymbol{K}_{ip}\Big)\boldsymbol{H}_i^{(1)}\Big)+\sum_{t=1}^{s}\sum_{i=1}^{N_{(t)}}\boldsymbol{\gamma}_i^{(t)}\\&\operatorname{tr}\Big(\boldsymbol{H}_i^{(t-1)\mathrm{T}}\Big(\sum_{j=1}^{N_{t-1}}dr_{ij}^{(t-1)}\boldsymbol{w}_{ij}^{(t-1)}\boldsymbol{H}_j^{(t-1)}\boldsymbol{H}_j^{(t-1)\mathrm{T}}\Big)\boldsymbol{H}_i^{(t-1)}\Big)+\\&\sum_{i=1}^{N_s}\boldsymbol{\beta}_i\operatorname{tr}(\boldsymbol{H}^{\mathrm{T}}(\boldsymbol{H}_i^{(s)}\boldsymbol{H}_i^{(s)\mathrm{T}})\boldsymbol{H})\\&\text{s.t. }\boldsymbol{H}^{\mathrm{T}}\boldsymbol{H}=\mathbf{I}_k,\boldsymbol{H}\in\Re^{n\times k},\boldsymbol{H}_i^{(t)\mathrm{T}}\boldsymbol{H}_i^{(t)}=\mathbf{I}_{c_t},\boldsymbol{H}_i^{(t)}\in\Re^{n\times c_t}\\&\qquad\boldsymbol{w}_i^{(t)\mathrm{T}}\boldsymbol{w}_i^{(t)}=1,\boldsymbol{w}_i^{(t)}\in\Re^{N_t},\boldsymbol{w}_i^{(1)\mathrm{T}}\boldsymbol{w}_i^{(1)}=1,\boldsymbol{w}_i^{(1)}\in\Re^{m}\\&\qquad\boldsymbol{w}_{ip}^{(1)},w_{ij}^{(t)}\in[0,1],\boldsymbol{\gamma}^{(t)\mathrm{T}}\boldsymbol{\gamma}^{(t)}=1,\boldsymbol{\gamma}_i^{(t)}\geqslant 0,\gamma\in\Re^{N_t}\\&\qquad\boldsymbol{dr}_{ip}^{(1)},dr_{ij}^{(t)}\in\{0,1\},\boldsymbol{\beta}^{\mathrm{T}}\boldsymbol{\beta}=1,\beta_j\geqslant 0,\beta\in\Re^{N_s}\\&\qquad n>c_1>c_2>\cdots>c_s>k\end{aligned}\tag{7}$$

其中每一层之间的平衡系数($\boldsymbol{w},\boldsymbol{\gamma},\boldsymbol{\beta}$)都添加了求和为 1 的约束以保证其合理性,$n>c_1>c_2>\cdots>k$ 表示每一层的聚类数是呈严格的递进关系,以此来保证信息蒸馏的理念。

### 2.2 SCHMKKM 算法的求解

式(7)需要优化 5 个变量,分别为 $\boldsymbol{H}$,$\boldsymbol{H}_i^{(t)}$,$\boldsymbol{\beta}$,$\boldsymbol{\gamma}$ 和 $\boldsymbol{w}$。考虑到所有变量的优化问题都是一个非凸的优化问题,优化困难,当优化其中一个变量而固定其他变量时,式(7)就变成了一个凸优化问题,因此我们采用逐个变量交替更新的优化策略,具体如下:

(1)更新 $\boldsymbol{H}$,固定其他变量,可以得到以下求解模型:

$$\begin{aligned}&\max_{\boldsymbol{H},\boldsymbol{\beta}}\operatorname{tr}\Big(\boldsymbol{H}^{\mathrm{T}}\sum_{i=1}^{N_s}\boldsymbol{\beta}_i\boldsymbol{H}_i^{(s)}\boldsymbol{H}_i^{(s)\mathrm{T}}\boldsymbol{H}\Big)\\&\text{s.t. }\boldsymbol{H}^{\mathrm{T}}\boldsymbol{H}=\mathbf{I}_k\end{aligned}\tag{8}$$

(2)更新 $\{\boldsymbol{H}_i^{(t)}\}_{t=1}^{s}$,在给出其他变量时可以得到以下求解模型:

$$\begin{aligned}&\max_{\boldsymbol{H}_i^{(t)}}\operatorname{tr}(\boldsymbol{H}_i^{(t)\mathrm{T}}\boldsymbol{A}_i^{(t)}\boldsymbol{H}_i^{(t)})\\&\text{s.t. }\boldsymbol{H}_i^{(t)\mathrm{T}}\boldsymbol{H}_i^{(t)}=\mathbf{I}_{c_t}\end{aligned}\tag{9}$$

其中,$\boldsymbol{A}_i^{(t)}$ 的取值有 3 种情况:

$$\boldsymbol{A}_i^{(t)}=\begin{cases}\sum_{j=1}^{N_{t-1}}dr_{ij}^{(t-1)}w_{ij}^{(t-1)}\boldsymbol{H}_j^{(t-1)}\boldsymbol{H}_j^{(t-1)\mathrm{T}}+\boldsymbol{\beta}_i\boldsymbol{H}\boldsymbol{H}^{\mathrm{T}}, & t=s\\ \boldsymbol{\gamma}_i^{(t-1)}\sum_{j=1}^{N_{t-1}}dr_{ij}^{(t-1)}\boldsymbol{w}_{ij}^{(t-1)}\boldsymbol{H}_j^{(t-1)}\boldsymbol{H}_j^{(t-1)\mathrm{T}}+\\ \boldsymbol{\gamma}_i^{(t+1)}\sum_{j=1}^{N_{t+1}}dr_{ij}^{(t+1)}\boldsymbol{w}_{ij}^{(t+1)}\boldsymbol{H}_j^{(t+1)}\boldsymbol{H}_j^{(t+1)\mathrm{T}}, & 1<t<s\\ \boldsymbol{\gamma}_i^{(t)}\sum_{j=1}^{N_t}dr_{ip}^{(t)}w_{ip}^{(t)}\boldsymbol{K}_P+\boldsymbol{\gamma}_i^{(t+1)}\sum_{j=1}^{N_{t+1}}dr_{ij}^{(t+1)}w_{ij}^{(t+1)}\\ \boldsymbol{H}_j^{(t+1)}\boldsymbol{H}_j^{(t+1)\mathrm{T}}, & t=1\end{cases}\tag{10}$$

(3)更新 $\boldsymbol{\beta}$,固定其他变量可以得到以下求解模型:

$$\begin{aligned}&\max_{\boldsymbol{\beta}}\boldsymbol{v}^{\mathrm{T}}\boldsymbol{\beta}\\&\text{s.t. }\boldsymbol{\beta}^{\mathrm{T}}\boldsymbol{\beta}=1,\boldsymbol{\beta}_i\geqslant 0\end{aligned}\tag{11}$$

其中 $\boldsymbol{v}$ 为:

$$\boldsymbol{v}_i=\operatorname{tr}(\boldsymbol{H}^{\mathrm{T}}\boldsymbol{H}_i^{(s)}\boldsymbol{H}_i^{(s)\mathrm{T}}\boldsymbol{H})\tag{12}$$

(4)更新 $\{\boldsymbol{\gamma}_i^{(t)}\}_{t=1}^{s}$,由于在更新 $\gamma$ 时,原始核 $\boldsymbol{K}$ 和隐藏层中的 $\boldsymbol{H}$ 以及其对应的 $\boldsymbol{dr}$ 和 $\boldsymbol{w}$ 矩阵均已得到,故可以通过以下求解模型更新 $\gamma$:

$$\begin{aligned}&\max_{\gamma}\boldsymbol{v}^{(t)\mathrm{T}}\boldsymbol{\gamma}^{(t)}\\&\text{s.t. }\boldsymbol{\gamma}^{\mathrm{T}}\boldsymbol{\gamma}=1,\boldsymbol{\gamma}_i\geqslant 0\end{aligned}\tag{13}$$

其中 $\boldsymbol{v}$ 为:

$$\begin{aligned}&\boldsymbol{v}_i^{(1)}=\operatorname{tr}\Big(\boldsymbol{H}_i^{(1)\mathrm{T}}\sum_{p=1}^{N_1}\boldsymbol{dr}_{ip}^{(1)}\boldsymbol{w}_{ip}^{(1)}\boldsymbol{K}_p\boldsymbol{H}_i^{(1)}\Big)\\&\boldsymbol{v}_i^{(t)}=\operatorname{tr}\Big(\boldsymbol{H}_i^{(t)\mathrm{T}}\Big(\sum_{j=1}^{N_{t-1}}\boldsymbol{dr}_{ij}^{(t-1)}\boldsymbol{w}_{ij}^{(t-1)}\boldsymbol{H}_j^{(t-1)}\boldsymbol{H}_j^{(t-1)\mathrm{T}}\Big)\boldsymbol{H}_i^{(t)}\Big)\end{aligned}\tag{14}$$

（5）更新 $\{w_i^{(t)}\}_{t=1}^{s}$，固定其他变量可以得到以下求解模型：

$$\max_{w} \boldsymbol{v}^{(t)\mathrm{T}} \boldsymbol{w}^{(t)} \quad \text{s.t. } \boldsymbol{w}_i^{(t)\mathrm{T}} \boldsymbol{w}_i^{(t)}=1, \boldsymbol{w}_{ij}^{(t)} \geqslant 0 \tag{15}$$

其中 $\boldsymbol{v}$ 为：

$$\boldsymbol{v}_i^{(1)}=\mathrm{tr}(\boldsymbol{\gamma}_i^{(1)} \boldsymbol{H}_i^{(1)\mathrm{T}}(\sum_{p=1}^{N_1} \boldsymbol{dr}_{ip}^{(1)} \boldsymbol{K}_p) \boldsymbol{H}_i^{(1)})$$

$$\boldsymbol{v}_i^{(t)}=\mathrm{tr}(\boldsymbol{\gamma}_i^{(t)} \boldsymbol{H}_i^{(t)\mathrm{T}}(\sum_{j=1}^{N_{t-1}} \boldsymbol{dr}_{ij}^{(t-1)} \boldsymbol{H}_j^{(t-1)} \boldsymbol{H}_j^{(t-1)\mathrm{T}}) \boldsymbol{H}_i^{(t)}) \tag{16}$$

对式（8）－式（9）可以归为一种统一的求解形式，如下：

$$\max_{U} \mathrm{tr}(\boldsymbol{U}^{\mathrm{T}} \boldsymbol{G} \boldsymbol{U}) \quad \text{s.t. } \boldsymbol{U}^{\mathrm{T}} \boldsymbol{U}=\mathbf{I}_c \tag{17}$$

式（15）是一个典型的 SVD 求解问题，可以通过对 $G$ 进行奇异值分解得到 $\boldsymbol{G}=\boldsymbol{Y}_c \boldsymbol{\Sigma}_c \boldsymbol{Q}_c^{\mathrm{T}}$，其中 $G$ 在式（8）中为 $\sum_{i=1}^{N_s} \boldsymbol{\beta}_i \boldsymbol{H}_i^{(s)} \boldsymbol{H}_i^{(s)\mathrm{T}}$，在式（9）中分别为对应不同隐藏层的 $\boldsymbol{A}_i^{(t)}$，$\{\boldsymbol{Y}_c, \boldsymbol{Q}_c\} \in \Re^{n \times c}$，$\boldsymbol{\Sigma}_c \in \Re^{c \times c}$。因此变量 $\boldsymbol{U}$ 的解为 $\boldsymbol{U}=\boldsymbol{Y}_c \boldsymbol{Q}_c^{\mathrm{T}}$。对于式（11）、式（13）、式（15）的解等同于如下模型的逐层求解：

$$\max_{\alpha} \boldsymbol{v}^{\mathrm{T}} \boldsymbol{\alpha} \quad \text{s.t. } \boldsymbol{\alpha}^{\mathrm{T}} \boldsymbol{\alpha}=1, \boldsymbol{\alpha}_i \geqslant 0 \tag{18}$$

式（17）的解为如下形式：

$$\boldsymbol{\alpha}_i=\frac{\boldsymbol{v}_i}{(\sum_{i=1}^{N_t} \boldsymbol{\alpha}_i^2)^{\frac{1}{2}}}, t=1,2,\cdots,s \tag{19}$$

其中，$\boldsymbol{v}$ 分别为式（12）、式（14）、式（16）。需要注意的是，模型的优化更新方式分为正向优化和反向优化。在优化过程中，与 HMKC[10] 的优化方式不同，在每一层隐藏层中更新时需考虑前后两层中所有结点与待更新结点之间的交互信息。SCHMKKM 优化过程如算法 1 所示。

**算法 1　稀疏连接的层次化多核 $K$-Means 算法（SCHMKKM）**

输入：预定义核 $\{\mathbf{K}_p\}_{p=1}^{m}$，聚类数 k，隐藏层大小 $\{c_t, N_t\}_{t=1}^{s}$，层数 s

输出：最终一致性划分矩阵 $\mathbf{H}$

1. 初始化 $\beta_i=1/\sqrt{N_s}$，$\boldsymbol{\beta} \in \Re^{N_s \times 1}$
2. 初始化 $\gamma_i^{(t)}=1/\sqrt{N_t}$，$\boldsymbol{\gamma}^{(t)} \in \Re^{N_t \times 1}$，$t=1,\cdots,s$
3. 初始化 $\mathbf{w}_{ij}^{(t)}=1/\sqrt{N_{t-1}}$，$\mathrm{w}_i^{(t)} \in \Re^{N_{t-1} \times 1}$，$j=1,\cdots,N_t$，$t=1,\cdots,s$
4. while not do convergence
5. 　使用 SVD 求解式（8）更新 $\mathbf{H}$
6. 　使用 SVD 求解式（9）更新 $\{\mathbf{H}_i^{(t)}\}_{t=1}^{s}$
7. 　结合式（12）和（19）求解式（11）更新 $\boldsymbol{\beta}$
8. 　结合式（14）和式（19）求解式（13）更新 $\boldsymbol{\gamma}$
9. 　结合式（16）和式（19）求解式（15）更新 $\mathbf{w}$
10. 判断收敛 $\left|\frac{F(t-1)-F(t)}{F(t)}\right| \leqslant 10^{-5}$，F 为式（7）的目标函数
11. end while.

### 2.3　复杂度和收敛性分析

更新式（8）时，使用 SVD 求解器复杂度为 $O(n^3)$；更新式（9）时，使用 SVD 求解器复杂度为 $\max(N_1, N_2, N_s) * O(n^3)$；更新式（11）时计算矩阵乘法的复杂度为 $O(n)$；更新式（13）和式（15）时计算矩阵乘法的复杂度为 $\max(N_1, N_2, N_s) * O(n^2)$。由于 $N_t \ll n$，因此 SCHMKKM 算法的复杂度为 $O(n^3)$。

当优化一个变量而固定其他变量时，式（8）、式（9）、式（11）、式（13）、式（15）的目标函数值都是单调递增的，且由于变量的正交约束和归一化约束可以保证目标函数有上界，因此模型（7）的目标函数值是收敛的。

## 3　实验与结果

### 3.1　对比方法和数据集

数据集具体细节如表 1 所列。在数据集上与 MKC 密切相关的 9 种算法以及全连接的层次化多核 $K$-Means 算法（FCHMKKM）进行对比。对比方法简介如下：

（1）Multiple Kernel $K$-Means Clustering with Matrix-induced Regularization（MKKM-MR，2016）[16]，一种使用矩阵诱导正则的多核 $K$-Means 聚类算法。

（2）Multiple Kernel Clustering with Local Kernel Alignment maximization（LKAMKC，2016）[17]，一种寻找局部最大一致性核的多核聚类算法。

（3）Optimal Neighborhood Kernel Clustering（ONKC，2017）[11]，一种寻找最优核的最优邻域核聚类算法。

（4）Low-rank Kernel learning for Graph matrix（LKGr，2018）[18]，一种将图约束与核学习统一的算法框架，目的是学习一个低秩核。

（5）Multiview Consensus Graph Clustering（MCGC，2019）[19]，一种构建一致性多视图的聚类算法。

（6）Joint Robust Multiple Kernel Subspace Clustering（JMKSC，2019）[20]，一种将熵度量权重和块对角正则优化联合的鲁棒多核子空间聚类算法。

（7）Optimal Neighborhood MKC with Adaptive Local Kernels（ON-ALK，2021）[12]，一种自适应局部核的最优邻域多核聚类算法。

表 1　对比实验的数据集

Table 1　Comparison of experimental datasets

| No. | Datasets | Samples | Features | Classes | Kernels |
|---|---|---|---|---|---|
| D1 | PROSTATE_GE | 102 | 5966 | 2 | 12 |
| D2 | CLL_SUB | 111 | 11340 | 3 | 12 |
| D3 | AR10P | 130 | 2400 | 10 | 12 |
| D4 | IONOSPHERE | 351 | 34 | 2 | 12 |
| D5 | WDBC | 569 | 30 | 2 | 12 |
| D6 | AUSTRALIAN | 690 | 14 | 2 | 12 |
| D7 | VEHICLE | 846 | 18 | 4 | 12 |
| D8 | PIE_POSE27 | 1428 | 1024 | 68 | 12 |
| D9 | YALEB | 2414 | 1024 | 38 | 12 |

（8）Projective Multiple Kernel Subspace Clustering（PMKSC，2021）[9]，一种利用低维分割空间中构造的多核图融合互补相似图的投影聚类算法。

（9）Hierarchical Multiple Kernel Clustering（HMKC，2021）[10]，一种最新的层次化多核聚类算法。

### 3.2　实验设置

以上 9 种对比方法的源码可以通过作者网站或相关网站获取，为了实验的公平性，本文均采用作者提供的源码进行对比。需要说明的是，以上方法均涉及超参数的设置以及采取网格搜索的策略获取最优结果，具体对比方法的超参数和本文实验设置如下：

(1)聚类中簇的个数采用的是真实数据集中给出的聚类数。

(2)所有对比方法的超参数设置如下：MKKMMR 的超参数搜索范围为$\{2^{-15},2^{-14},\cdots,2^{15}\}$；LKAMKC 的两个超参数搜索范围为$\{2^{-15},2^{-14},\cdots,2^{15}\}$和$\{0.1,0.2,\cdots,1\}$；ONKC 的两个超参数搜索范围都为$\{2^{-15},2^{-13},\cdots,2^{13},2^{15}\}$；LKGr 的前 2 个超参数范围为$\{10^{-5},10^{-3},\cdots,10^{3},10^{5}\}$，后 2 个超参数值为$10^{-2}$；JMKSC 的 3 个超参数搜索范围分别为$\{10^{-4},10^{-3},\cdots,1,10\}$，$\{1,5,10,\cdots,30\}$和$\{0.1,1,5,10,\cdots,30\}$；MCGC 邻域超参数搜索范围为$\{3,5,7,9,11,13,15\}$；ONALK 的两个超参数搜索范围为$\{-0.5,-0.4,\cdots,0.4,0.5\}$和$\{2^{-15},2^{-14},\cdots,2^{15}\}$；PMKSC 的 3 个超参数搜索中前 2 个的范围为$\{2^{-5},2^{-3},\cdots,2^{3},2^{5}\}$，最后 1 个的范围为$\{10^{-5},10^{-3},\cdots,10^{3},10^{5}\}$；HMKC 两个隐藏层的聚类个数搜索范围为$\{3k,4k,\cdots,10k\}$和$\{2k,3k,4k,\cdots,10k\}$。

由于计算资源的限制，本文同 HMKC 一样设置两层隐藏层进行实验，因此 SCHMKKM 算法的参数为 $c_1=\{3k,4k,\cdots,10k\}$，$c_2=\{2k,3k,4k,5k\}$，其中 $c_1>c_2$ 稀疏率设置为$\{0.5,0.6,0.7,0.8\}$，隐藏层的结点数设置为 $N_1=\{8,10,12,14\}$，$N_2=\{6,8,10\}$；FCHMKKM 算法的稀疏率设置为 0，其他参数设置同 SCHMKKM 一样。

(3)实验中所有算法采用的核函数包括 7 个 RBF 核 $K(\boldsymbol{x}_i,\boldsymbol{x}_j)=\exp(-\|\boldsymbol{x}_i-\boldsymbol{x}_j\|/2(\varepsilon*d)^2)$，其中 $d$ 为样本间最大欧氏距离，方差 $\varepsilon$ 的取值为$\{2^{-3},\cdots,2^{3}\}$，4 个多项式核 $K(\boldsymbol{x}_i,\boldsymbol{x}_j)=(a+\boldsymbol{x}_i^{\mathrm{T}}\boldsymbol{x}_j)^b$，其中 $a$ 的取值范围为$\{0,1\}$，$b$ 的取值范围为$\{2,4\}$，1 个线性核 $K(\boldsymbol{x}_i,\boldsymbol{x}_j)=\boldsymbol{x}_i^{\mathrm{T}}\boldsymbol{x}_j$。

(4)以上对比方法除 MCGC 外均需要进行后处理来获得离散化的聚类结果，故本文采用 $K$-Means 进行离散化，MCGC 采用切割连通图的方式获取聚类结果。为了消除初始化带来的影响，所有算法均运行 30 次，取网格搜索得到的结果的平均值及其方差。

(5)本文所提出的算法源码[1]已公开。

### 3.3 实验结果与分析

在对比实验中，我们采用 3 种常见的聚类度量方法作为评价指标，分别为聚类精度(Accuracy，ACC)、归一化互信息(Normalized Mutual Information，NMI)、兰德系数(Rand Index，RI)。结果如表 2－表 4 所列，最佳结果用黑体表示，次优结果用下划线表示。

表 2 ACC 实验结果对比

Table 2 Comparison of ACC experimental results

(单位：%)

| Algs\Ds | D1 | D2 | D3 | D4 | D5 | D6 | D7 | D8 | D9 | Avg |
|---|---|---|---|---|---|---|---|---|---|---|
| MKKM-MR (2016) | 56.93±0.01 | 50.63±0.00 | 42.15±0.30 | 72.08±0.00 | 88.58±0.00 | 69.13±0.00 | 45.06±0.02 | 83.22±0.10 | 26.06±0.02 | 59.32 |
| LKAMKC (2016) | 64.71±0.00 | 60.60±0.01 | 42.10±0.00 | 72.93±0.00 | 90.86±0.00 | 69.13±0.00 | 45.74±0.00 | 35.99±0.00 | 9.69±0.00 | 54.64 |
| ONKC(2017) | 57.61±0.02 | 50.45±0.07 | 42.41±0.42 | 73.22±0.00 | 89.03±0.03 | 69.42±0.00 | 44.95±0.01 | 84.13±0.08 | 25.34±0.02 | 59.62 |
| LKGr(2018) | 63.73±0.00 | 52.19±0.03 | 10.77±0.00 | 71.04±0.00 | 68.37±0.00 | 59.71±0.00 | 43.21±0.01 | 9.84±0.00 | 7.20±0.00 | 42.90 |
| JMKSC(2019) | 60.46±0.00 | 54.05±0.00 | 38.05±0.10 | 65.31±0.05 | 79.94±0.00 | 58.70±0.00 | 41.49±0.00 | 76.46±0.12 | 5.83±0.00 | 53.37 |
| MCGC(2019) | 63.73±0.00 | 42.34±0.00 | 44.62±0.00 | 65.53±0.00 | 87.52±0.00 | 68.99±0.00 | 25.89±0.00 | 72.13±0.00 | 34.01±0.00 | 56.08 |
| ON-ALK(2021) | 74.02±0.00 | 50.45±0.00 | 36.82±0.16 | 74.36±0.00 | 85.20±0.77 | 69.57±0.00 | 46.12±0.00 | 78.03±0.20 | 22.25±0.02 | 59.65 |
| PMKSC(2021) | 64.71±0.00 | 49.22±0.03 | 38.26±0.27 | 70.37±0.00 | 91.92±0.00 | **69.71±0.00** | 44.44±0.00 | 83.42±0.09 | 26.40±0.02 | 59.83 |
| HMKC(2021) | 63.79±0.37 | 48.32±0.08 | 46.56±0.23 | 71.79±0.00 | 92.09±0.00 | 67.35±0.00 | 41.92±0.03 | 84.72±0.09 | 29.33±0.04 | 60.65 |
| FCHMKKM | 85.29±0.00 | 58.83±0.00 | 42.85±0.04 | 87.46±0.00 | 90.51±0.00 | 68.41±0.00 | 45.51±0.00 | 47.25±0.02 | 32.12±0.01 | 62.03 |
| SCHMKKM | **86.18±0.03** | **63.51±0.02** | **56.38±0.13** | **90.03±0.03** | **93.20±0.00** | 69.55±0.00 | **53.68±0.12** | **84.93±0.12** | **40.06±0.02** | 70.84 |

表 3 NMI 实验结果对比

Table 3 Comparison of NMI experimental results

(单位：%)

| Algs\Ds | D1 | D2 | D3 | D4 | D5 | D6 | D7 | D8 | D9 | Avg |
|---|---|---|---|---|---|---|---|---|---|---|
| MKKM-MR (2016) | 1.39±0.00 | 14.71±0.01 | 42.27±0.26 | 22.27±0.00 | 48.70±0.00 | 10.27±0.00 | 17.47±0.02 | 95.48±0.01 | 42.05±0.01 | 32.73 |
| LKAMKC (2016) | 6.86±0.00 | **25.76±0.00** | 42.12±0.26 | 21.44±0.00 | 54.29±0.00 | 10.29±0.00 | 19.36±0.00 | 69.58±0.00 | 14.73±0.00 | 29.38 |
| ONKC(2017) | 1.70±0.00 | 14.82±0.09 | 42.42±0.45 | 24.11±0.00 | 50.00±0.11 | 10.61±0.00 | 18.94±0.02 | 95.93±0.01 | 41.46±0.01 | 33.33 |
| LKGr(2018) | 6.73±0.00 | 19.10±0.05 | 8.25±0.00 | 11.59±0.00 | 11.37±0.00 | 2.27±0.00 | 15.74±0.02 | 31.50±0.00 | 9.03±0.00 | 12.84 |
| JMKSC(2019) | 3.61±0.00 | 18.69±0.00 | 37.83±0.12 | 6.93±0.02 | 30.57±0.00 | 3.27±0.00 | 15.58±0.00 | 91.77±0.01 | 6.38±0.00 | 23.85 |
| MCGC(2019) | 6.47±0.00 | 2.28±0.00 | 41.27±0.00 | 9.35±0.00 | 47.76±0.00 | 9.98±0.00 | 0.37±0.00 | 89.80±0.00 | 46.80±0.00 | 28.23 |
| ON-ALK(2021) | 17.35±0.01 | 14.47±0.00 | 34.62±0.19 | 20.37±0.00 | 42.51±02.75 | 10.76±0.00 | 18.91±0.00 | 92.13±0.03 | 35.36±0.03 | 31.83 |
| PMKSC(2021) | 6.60±0.00 | 10.56±0.05 | 38.84±0.22 | 13.42±0.00 | 57.86±0.00 | **10.94±0.00** | 16.96±0.00 | 95.73±0.01 | 42.51±0.00 | 32.60 |
| HMKC(2021) | 6.75±0.24 | 13.22±0.14 | 43.05±0.15 | 15.08±0.00 | 58.48±0.00 | 9.13±0.00 | 18.52±0.05 | 96.13±0.01 | 45.60±0.01 | 34.00 |
| FCHMKKM | 40.06±0.00 | 15.94±0.00 | 39.77±0.03 | 44.43±0.00 | 56.41±0.00 | 10.37±0.00 | 15.32±0.00 | 76.17±0.01 | 44.10±0.00 | 38.06 |
| SCHMKKM | **43.26±0.24** | 24.61±0.06 | **54.76±0.10** | **51.29±0.34** | **62.65±0.00** | 10.72±0.00 | **27.28±0.02** | **96.22±0.01** | **51.76±0.01** | **46.95** |

[1] https://gitee.com/wl575264909/schmkkm.git

**表 4　RI 实验结果对比**

Table 4　Comparison of RI experimental results

（单位：%）

| Algs\Ds | D1 | D2 | D3 | D4 | D5 | D6 | D7 | D8 | D9 | Avg |
|---|---|---|---|---|---|---|---|---|---|---|
| MKKM-MR (2016) | 50.49±0.00 | 56.10±0.00 | 85.61±0.01 | 59.64±0.00 | 79.73±0.00 | 57.26±0.00 | 67.24±0.02 | 99.52±0.00 | 95.55±0.00 | 72.35 |
| LKAMKC (2016) | 53.87±0.00 | 59.65±0.00 | 85.75±0.01 | 60.41±0.00 | 83.36±0.00 | 57.26±0.00 | 67.66±0.00 | 97.82±0.00 | 94.72±0.00 | 73.39 |
| ONKC(2017) | 50.71±0.00 | 55.57±0.01 | 85.81±0.01 | 60.67±0.00 | 80.49±0.07 | 57.48±0.00 | 67.17±0.00 | 99.56±0.00 | 95.53±0.00 | 72.55 |
| LKGr(2018) | 53.31±0.00 | 52.69±0.02 | 21.00±0.00 | 58.73±0.00 | 56.67±0.00 | 51.82±0.00 | 64.64±0.03 | 97.10±0.00 | 94.65±0.00 | 61.18 |
| JMKSC(2019) | 51.72±0.00 | 52.45±0.00 | 81.17±0.07 | 55.05±0.03 | 67.87±0.00 | 51.44±0.00 | 61.85±0.00 | 99.17±0.00 | 68.06±0.00 | 65.42 |
| MCGC(2019) | 53.31±0.00 | 48.75±0.00 | 77.92±0.00 | 54.69±0.00 | 78.12±0.00 | 57.15±0.00 | 25.32±0.00 | 98.47±0.00 | 95.12±0.00 | 65.43 |
| ON-ALK(2021) | 61.16±0.00 | 56.15±0.00 | 84.32±0.01 | 61.76±0.00 | 76.28±0.94 | 57.59±0.00 | 66.76±0.00 | 99.29±0.00 | 95.34±0.00 | 73.18 |
| PMKSC(2021) | 53.87±0.00 | 54.73±0.02 | 84.87±0.01 | 58.18±0.00 | 85.11±0.00 | **57.71±0.00** | 67.38±0.00 | 99.54±0.00 | 95.54±0.00 | 72.99 |
| HMKC(2021) | 54.10±0.11 | 55.83±0.01 | 85.87±0.01 | 59.38±0.00 | 85.41±0.00 | 55.96±0.00 | 66.41±0.00 | **99.58±0.00** | 95.64±0.00 | 73.13 |
| FCHMKKM | 74.67±0.00 | 59.32±0.00 | 85.74±0.00 | 78.01±0.00 | 82.79±0.00 | 56.71±0.00 | 67.81±0.00 | 98.13±0.00 | 95.63±0.00 | 77.65 |
| SCHMKKM | **76.01±0.07** | **60.14±0.01** | **88.58±0.00** | **82.06±0.07** | **87.30±0.00** | 57.58±0.00 | **71.64±0.00** | 99.57±0.00 | **95.78±0.00** | **79.85** |

从表 2—表 4 可以得出以下结论：

（1）在几乎所有数据集上（除 AUSTRALIAN 数据集），SCHMKKM 在 3 个聚类指标上都取得了最好的结果。在其他数据集上 ACC 比 9 种对比方法中最好的结果平均高出 10.19%，NMI 平均高出 12.95%，RI 平均高出 6.46%。对比全连接策略的层次化多核聚类，本文方法在 ACC 上平均高出 8.81%，NMI 平均高出 8.89%，RI 平均高出 2.2%。

（2）在表 3 中，NMI 在 CLL_SUB 和 AUSTRALIAN 数据集上分别低于对比方法 0.15% 和 0.22%，数值较为接近；在其他数据集上都远远高出对比方法。

（3）在对比方法中，表现较好的方法大多考虑到了局部核融合的方法（LKAMKC 和 ONALK），HMKC 在对比方法中的表现也较为突出。这说明考虑层次化和局部核融合的方法对多核融合有较好的效果。

（4）从表 2—表 4 的最后一列结果可以看出：首先，考虑邻层结点之间信息交互的算法（FCHMKKM 和 SCHMKKM）要明显优于 HMKC 算法和其他对比方法；其次，稀疏连接的策略明显好于全连接的策略，原因可能是在每一次的正向和反向优化中，由于每一层的结点考虑的都是所有结点之间的融合，因此造成了每一层结点的差异性减小，结点之间相似度变高，最终一致性划分矩阵受限于最后一层的划分矩阵；最后，全连接策略的性能明显优于 HMKC 算法，说明考虑层与层之间结点的信息交互是有必要的。

（5）图 2 给出了本文提出的两种核融合策略在 PROSTATE_GE 数据集上的收敛情况，从图中可以看出目标函数是呈单调递增的，并且在较少的迭代次数下达到了收敛状态。

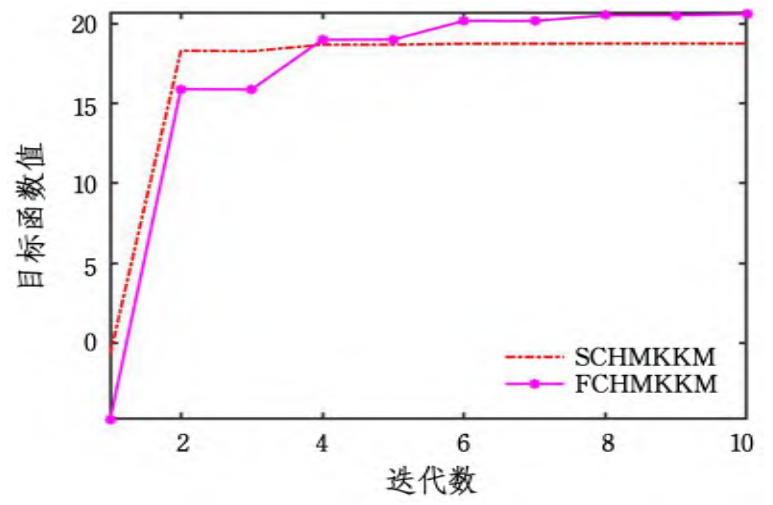

**图 2　SCHMKKM 和 FCHMKKM 目标函数收敛情况**

Fig. 2　Objective convergence of SCHMKKM and FCHMKKM

（6）图 3—图 5 给出了在 PROSTATE_GE 数据集上本文提出的算法对超参数聚类精度的敏感性情况。

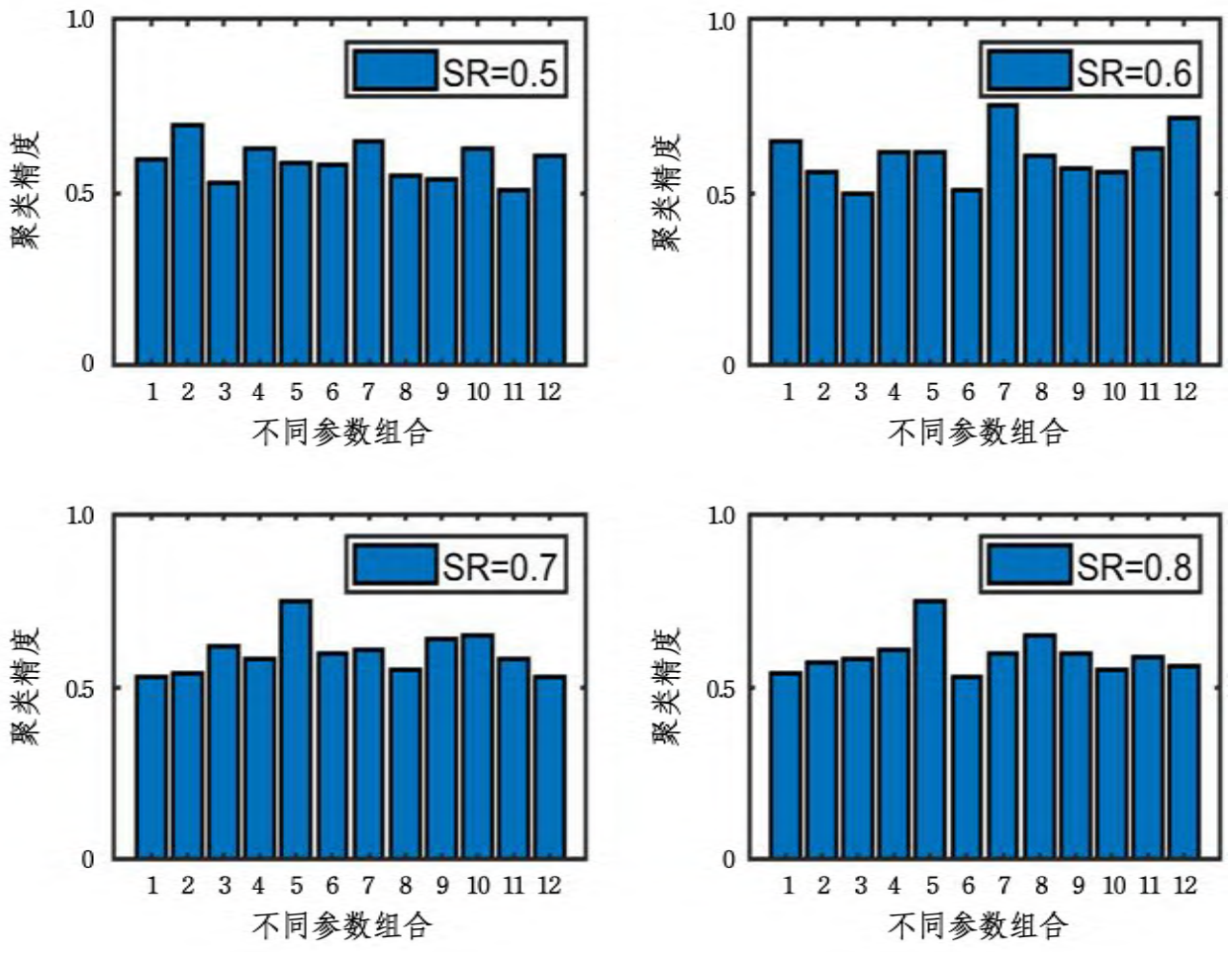

**图 3　不同稀疏率下对聚类精度的敏感性**

Fig. 3　Sensitivity to clustering accuracy at different sparsity rates

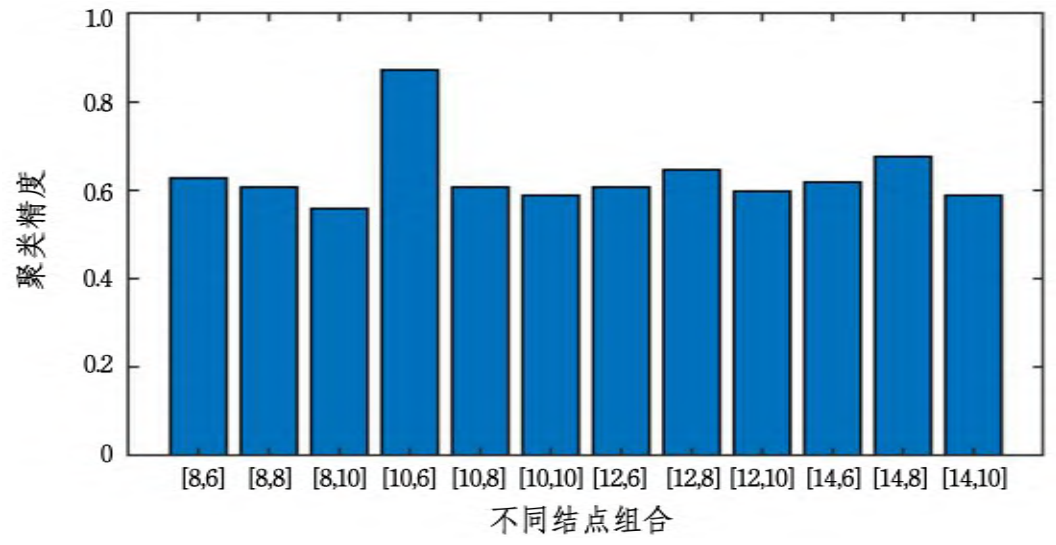

**图 4　不同结点数时对聚类精度的敏感性**

Fig. 4　Sensitivity to clustering accuracy with different number of nodes

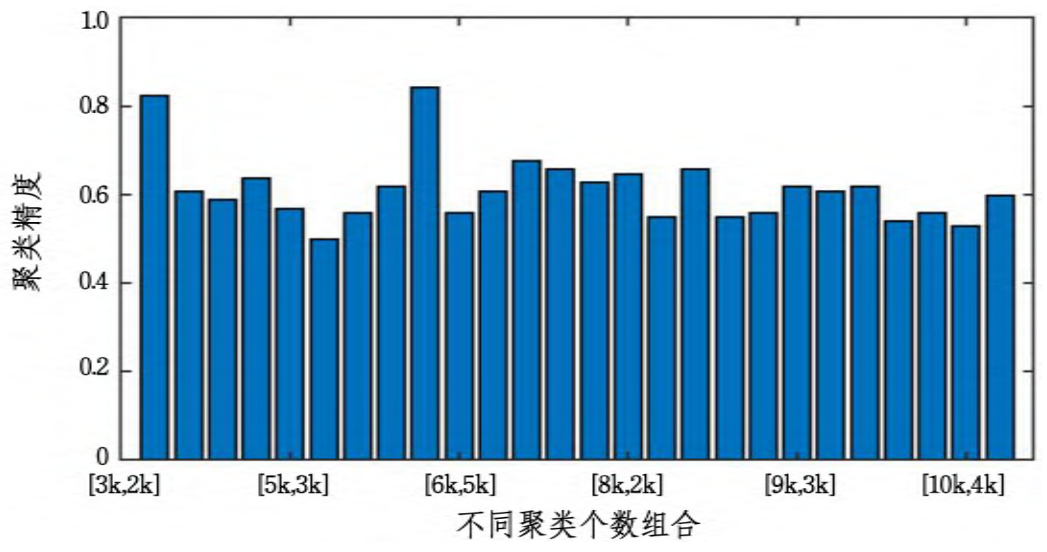

**图 5　不同聚类个数下对聚类精度的敏感性**

Fig. 5　Sensitivity to clustering accuracy with different numbers of clustering

在图 3 中,我们选取在固定稀疏率下所有参数组合(312 种)对应的聚类精度,并随机选取 12 组结果。从图中可以看出,本文方法对参数的敏感程度较高,在特定参数组合下的聚类精度会明显高于大多数参数组合。在图 4 中,我们选取在 PROSTATE_GE 数据集上最优的一组参数,固定隐藏层的聚类个数和稀疏率($c_1$ 为 $6k$,$c_2$ 为 $3k$,稀疏率为 0.5),测试不同结点数对聚类精度的影响。从图中可以看出,在第一层结点数为 10、第二层结点数为 6 的情况下,聚类精度最高;在其他情况下,结点数对聚类的精度敏感性较低。在图 5 中,我们选取在 PROSTATE_GE 数据集上最优的一组参数并固定第一层结点数为 8,第二层结点数为 10,稀疏率为 0.6,测试不同聚类个数对聚类精度的影响。从图中可以看出,不同聚类数对聚类精度的影响较大,当两层之间的聚类个数相差较大时,聚类性能较好。从以上分析可以得出,SCHMKKM 算法的超参数对聚类性能的敏感性较大,对超参数的选择较为重要。

(7)为了更加明确超参数的选择,图 6 给出了 3 种参数组合在聚类精度和时间效率上的对比图,具体如下:选择层数为 2,隐藏层聚类数分别为 $c_1=\{3k,4k,\cdots,10k\}$,$c_2=\{2k,3k,4k,5k\}$,其中 $c_1>c_2$ 稀疏率为{0.5,0.6,0.7,0.8},隐藏层的结点数为 $N_1=\{8,10,12,14\}$,$N_2=\{6,8,10\}$。将其排列组合构成参数集合,并选取此参数集在 AR10P 数据集上进行实验,将其在不同参数组合下的聚类精度和时间归一化在同一个区间。可以看到,随着隐藏层簇大小、结点数以及稀疏率的增加,聚类精度反而呈现下降趋势,并且还会导致运行时间的增加。综合考虑到聚类精度和运行时间效率,我们进一步将参数设置如下:层数设置为 2,隐藏层聚类数分别为 $c_1=\{3k,4k,5k\}$,$c_2=\{2k,3k,4k\}$,其中 $c_1>c_2$ 稀疏率为{0.5,0.6,0.7,0.8},隐藏层的结点数为 $N_1=\{8,10,12,14\}$,$N_2=\{6,8,10\}$。

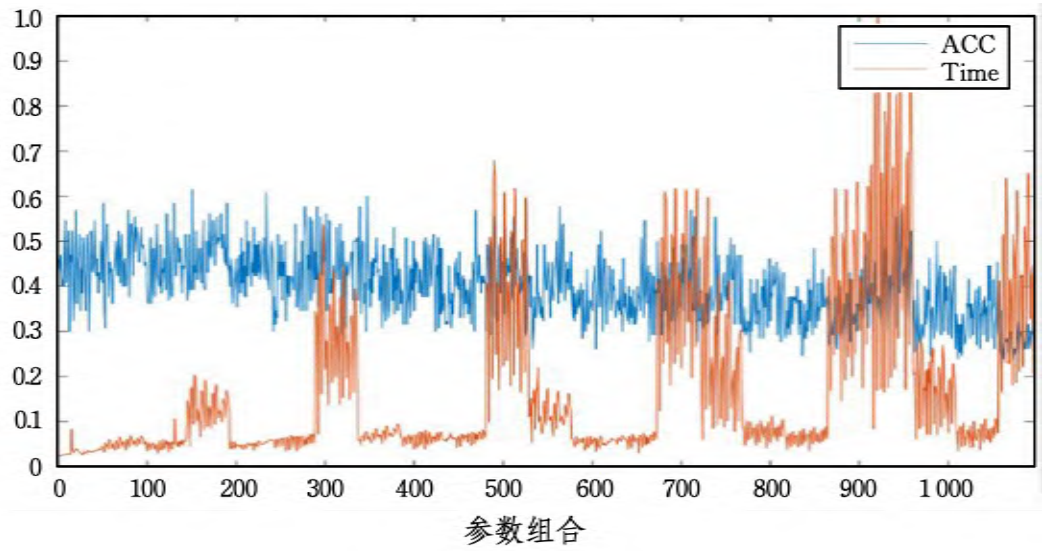

图 6 不同参数组合下的 ACC 与时间对比

Fig. 6 ACC and time comparison with different parameter combinations

(8)为了更加直观地展示全连接和稀疏连接策略对模型的影响,本文在 YALEB,PIE_POSE27 和 AR10P 数据集上展示了两者的 ACC 在不同参数组合下的聚类精度对比,如图 7 所示。从图中可以清晰地看到,稀疏连接策略相比全连接策略有着压倒性的优势。对后融合的多核聚类来说,影响最终一致性划分矩阵的聚类性能的因素包括每一层划分矩阵的多样性和其权重的影响,因此稀疏连接优于全连接策略可能有以下原因:1)全连接的策略可能会导致每一层的划分矩阵失去多样性,因为每一层的每一个划分矩阵在初始化时都是来自相同数据的线性组合,所以会导致每一层划分矩阵之间的相似性提高;2)从图 2 可以看到,全连接策略很快就能达到收敛状态,而稀疏连接策略呈阶梯状不断优化更新,意味着其权重和划分矩阵优化潜能大于全连接策略。综上所述,稀疏连接策略可以为每一层的划分矩阵增加多样性,间接导致最终的一致性划分矩阵多样性增加,进而学习到最优的一致性划分矩阵,提升了聚类性能。

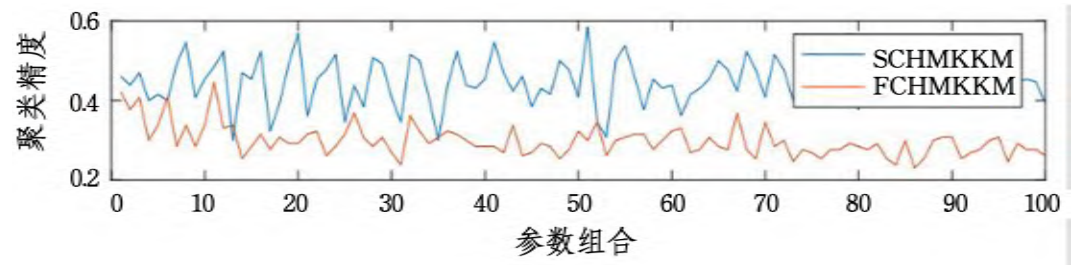

(a) AR10P

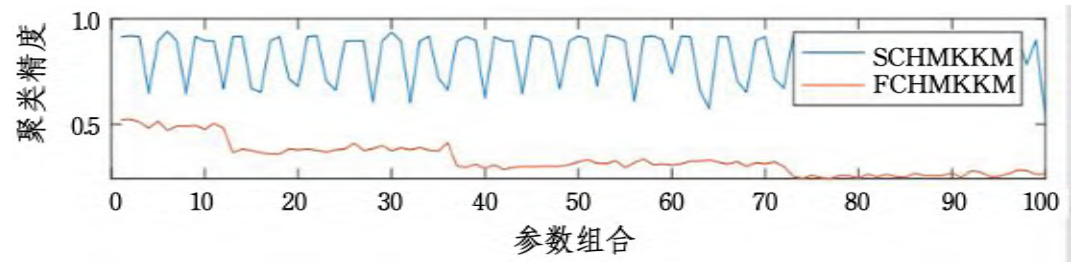

(b) PIEPOSE27

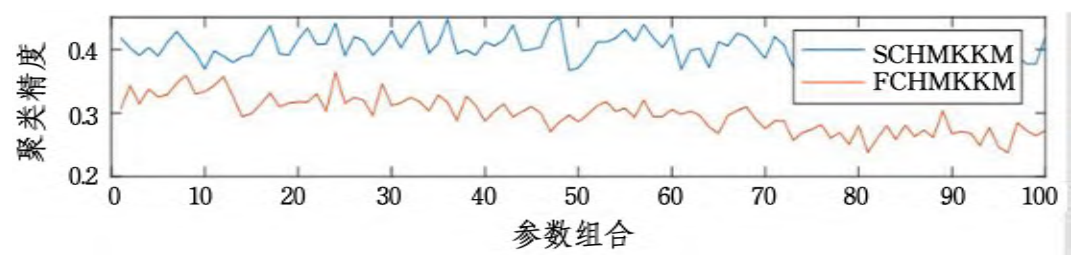

(c) YALEB

图 7 YALEB,PIEPOSE27 和 AR10P 数据集上的全连接策略和稀疏连接策略对比

Fig. 7 Comparison of full connection strategy and sparse connection strategy in datasets YALEB, PIEPOSE27 and AR10P

(9)表 5 列出了所有方法在 YALEB,PIE_POSE27 和 AR10P 数据集上的运行时间,具体为每种对比方法在所有参数组合上的平均运行时间。由于本文方法的模型比大多数对比方法更复杂,因此运行时间长于大部分对比方法。可以看到,随着数据量的增加,稀疏连接和全连接策略相比 HMKC 大幅度增加,其原因在于每一层结点数之间交互的线性时间增加。稀疏连接策略相比全连接策略运行时间较短,稀疏连接运行时间会随着稀疏率的降低逼近全连接策略。

表 5 时间对比

Table 5 Comparison of running times

(单位:s)

| Algs\Ds | AR10P | PIE_POSE27 | YALEB |
|---|---|---|---|
| MKKMMR | 0.04 | 33.02 | 139.42 |
| LKAMKC | 0.02 | 11.16 | 50.69 |
| ONKC | 0.11 | 20.31 | 55.06 |
| LKGr | 0.46 | 136.76 | 427.85 |
| JMKSC | 0.48 | 39.41 | 118.10 |
| MCGC | 0.90 | 31.66 | 251.06 |
| ONALK | 0.51 | 39.47 | 149.33 |
| PMKSC | 8.15 | 341.46 | 1512.44 |
| HMKC | 11.02 | 73.44 | 138.81 |
| FCHMKKM | 17.10 | 268.98 | 465.83 |
| SCHMKKM | 15.36 | 262.64 | 402.33 |

**结束语** 本文提出了一种新的基于稀疏连接与层次化聚类的多核聚类算法,并与多个最新的多核方法进行实验对比。SCHMKKM 算法在 3 个聚类指标上都优于 HMKC 与 FCHMKKM 算法,这从实验上证明了稀疏连接的层次化多核聚类算法比全连接的层次化多核聚类算法性能更优;也证明了在

层次化多核聚类算法中每一层考虑多个核函数的融合优于单核传递。但本文缺少理论依据证明稀疏连接的优越性，并且搜索参数偏多，这会导致算法的可移植性降低。在未来的工作中我们将会针对这两点缺陷对算法进行优化并期望从中衍生出更好的多核聚类算法。

## 参 考 文 献

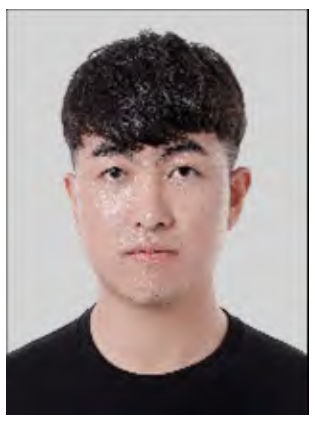

**WANG Lei**, born in 1996, postgraduate, is a member of China Computer Federation. His main research interests include data mining and multi-view clustering.

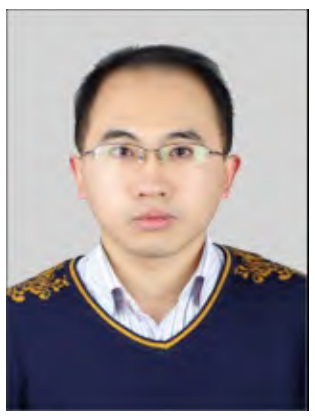

**DU Liang**, born in 1985, Ph. D, is a member of China Computer Federation. His main research interests include data mining, feature selection and clustering analysis.

（责任编辑：何杨）